\documentclass{article}

\usepackage[preprint]{spconf,amsmath,graphicx}
\copyrightnotice{\copyright\ IEEE 2018}
\toappear{To appear in {\it Proc.\ SLT2018,
            Dec 18-21, 2018, Athens, Greece}}

\usepackage{amsmath}
\usepackage{flexisym}
\usepackage{dcolumn}
\usepackage{kotex}
\usepackage{color}
\usepackage{hhline}
\usepackage{verbatim}
\usepackage{multirow}
\usepackage{amssymb}
\usepackage{rotating}
\usepackage{subfigure}
\usepackage{dblfloatfix}
\usepackage{enumitem}
\newcolumntype{L}[1]{>{\raggedright\let\newline\\\arraybackslash\hspace{0pt}}m{#1}}
\newcolumntype{C}[1]{>{\centering\let\newline\\\arraybackslash\hspace{0pt}}m{#1}}
\newcolumntype{R}[1]{>{\raggedleft\let\newline\\\arraybackslash\hspace{0pt}}m{#1}}

\newcommand\Tstrut{\rule{0pt}{2.5ex}}         

\title{Multimodal Speech Emotion Recognition Using Audio and Text}
\name{Seunghyun Yoon, Seokhyun Byun, and Kyomin Jung}
\address{
Dept. of Electrical and Computer Engineering, Seoul National University, Seoul, Korea \\ { \{mysmilesh,~byuns9334,~kjung\}@snu.ac.kr}
}

\begin{document}
%
\maketitle
\begin{abstract}


Speech emotion recognition is a challenging task, and extensive reliance has been placed on models that use audio features in building well-performing classifiers. In this paper, we propose a novel deep dual recurrent encoder model that utilizes text data and audio signals simultaneously to obtain a better understanding of speech data. As emotional dialogue is composed of sound and spoken content, our model encodes the information from audio and text sequences using dual recurrent neural networks (RNNs) and then combines the information from these sources to predict the emotion class. This architecture analyzes speech data from the signal level to the language level, and it thus utilizes the information within the data more comprehensively than models that focus on audio features. Extensive experiments are conducted to investigate the efficacy and properties of the proposed model. Our proposed model outperforms previous state-of-the-art methods in assigning data to one of four emotion categories (i.e., \textit{angry}, \textit{happy}, \textit{sad} and \textit{neutral}) when the model is applied to the IEMOCAP dataset, as reflected by accuracies ranging from 68.8\% to 71.8\%.

\end{abstract}
\section{Introduction}
\label{sec:intro}

Recently, deep learning algorithms have successfully addressed problems in various fields, such as image classification, machine translation, speech recognition, text-to-speech generation and other machine learning related areas~\cite{krizhevsky2012imagenet, bahdanau2014neural, amodei2016deep}. 
Similarly, substantial improvements in performance have been obtained when deep learning algorithms have been applied to statistical speech processing~\cite{graves2006connectionist}.
These fundamental improvements have led researchers to investigate additional topics related to human nature, which have long been objects of study. 
One such topic involves understanding human emotions and reflecting it through machine intelligence, such as emotional dialogue models~\cite{Zhou2018Emotional,huang2018automatic}.

In developing emotionally aware intelligence, the very first step is building robust emotion classifiers that display good performance regardless of the application; this outcome is considered to be one of the fundamental research goals in affective computing~\cite{busso2014toward}.
In particular, the speech emotion recognition task is one of the most important problems in the field of paralinguistics. This field has recently broadened its applications, as it is a crucial factor in optimal human-computer interactions, including dialog systems.
The goal of speech emotion recognition is to predict the emotional content of speech and to classify speech according to one of several labels (i.e., \textit{happy}, \textit{sad}, \textit{neutral}, and \textit{angry}). Various types of deep learning methods have been applied to increase the performance of emotion classifiers; however, this task is still considered to be challenging for several reasons. First, insufficient data for training complex neural network-based models are available, due to the costs associated with human involvement. 
Second, the characteristics of emotions must be learned from low-level speech signals. Feature-based models display limited skills when applied to this problem.

To overcome these limitations, we propose a model that uses high-level text transcription, as well as low-level audio signals, to utilize the information contained within low-resource datasets to a greater degree. Given recent improvements in automatic speech recognition (ASR) technology~\cite{yu2016automatic,amodei2016deep,GoogleCloudSpeechAPI,MicrosofSpeechAPI}, speech transcription can be carried out using audio signals with considerable skill. 
The emotional content of speech is clearly indicated by the emotion words contained in a sentence~\cite{xu2008constructing}, such as ``lovely'' and ``awesome,'' which carry strong emotions compared to generic (non-emotion) words, such as ``person'' and ``day.'' 
Thus, we hypothesize that the speech emotion recognition model will be benefit from the incorporation of high-level textual input.

In this paper, we propose a novel deep dual recurrent encoder model that simultaneously utilizes audio and text data in recognizing emotions from speech. Extensive experiments are conducted to investigate the efficacy and properties of the proposed model. Our proposed model outperforms previous state-of-the-art methods by 68.8\% to 71.8\% when applied to the IEMOCAP dataset, which is one of the most well-studied datasets. Based on an error analysis of the models, we show that our proposed model accurately identifies emotion classes. Moreover, the \textit{neutral} class misclassification bias frequently exhibited by previous models, which focus on audio features, is less pronounced in our model.

\section{Related work}
Classical machine learning algorithms, such as hidden Markov models (HMMs), support vector machines (SVMs), and decision tree-based methods, have been employed in speech emotion recognition problems~\cite{seehapoch2013speech, schuller2003hidden, lee2011emotion}. 
Recently, researchers have proposed various neural network-based architectures to improve the performance of speech emotion recognition. An initial study utilized deep neural networks (DNNs) to extract high-level features from raw audio data and demonstrated its effectiveness in speech emotion recognition~\cite{han2014speech}.
With the advancement of deep learning methods, more complex neural-based architectures have been proposed. Convolutional neural network (CNN)-based models have been trained on information derived from raw audio signals using spectrograms or audio features such as Mel-frequency cepstral coefficients (MFCCs) and low-level descriptors (LLDs)~\cite{bertero2017first, badshah2017speech, aldeneh2017using}.
These neural network-based models are combined to produce higher-complexity models~\cite{busso2008iemocap,satt2017efficient}, and these models achieved the best-recorded performance when applied to the IEMOCAP dataset.


Another line of research has focused on adopting variant machine learning techniques combined with neural network-based models. One researcher utilized the multiobject learning approach and used gender and naturalness as auxiliary tasks so that the neural network-based model learned more features from a given dataset~\cite{kim2017towards}.
Another researcher investigated transfer learning methods, leveraging external data from related domains ~\cite{gideon2017progressive}.

As emotional dialogue is composed of sound and spoken content, researchers have also investigated the combination of acoustic features and language information, built belief network-based methods of identifying emotional key phrases, and assessed the emotional salience of verbal cues from both phoneme sequences and words~\cite{schuller2004speech, gamage2017salience}.
However, none of these studies have utilized information from speech signals and text sequences simultaneously in an end-to-end learning neural network-based model to classify emotions.

\section{Model}
This section describes the methodologies that are applied to the speech emotion recognition task. We start by introducing the recurrent encoder model for the audio and text modalities individually. We then propose a multimodal approach that encodes both audio and textual information simultaneously via a dual recurrent encoder.

\subsection{Audio Recurrent Encoder (ARE)}
\begin{figure}[t]
\small
\centering
\includegraphics[width=1.0\columnwidth]{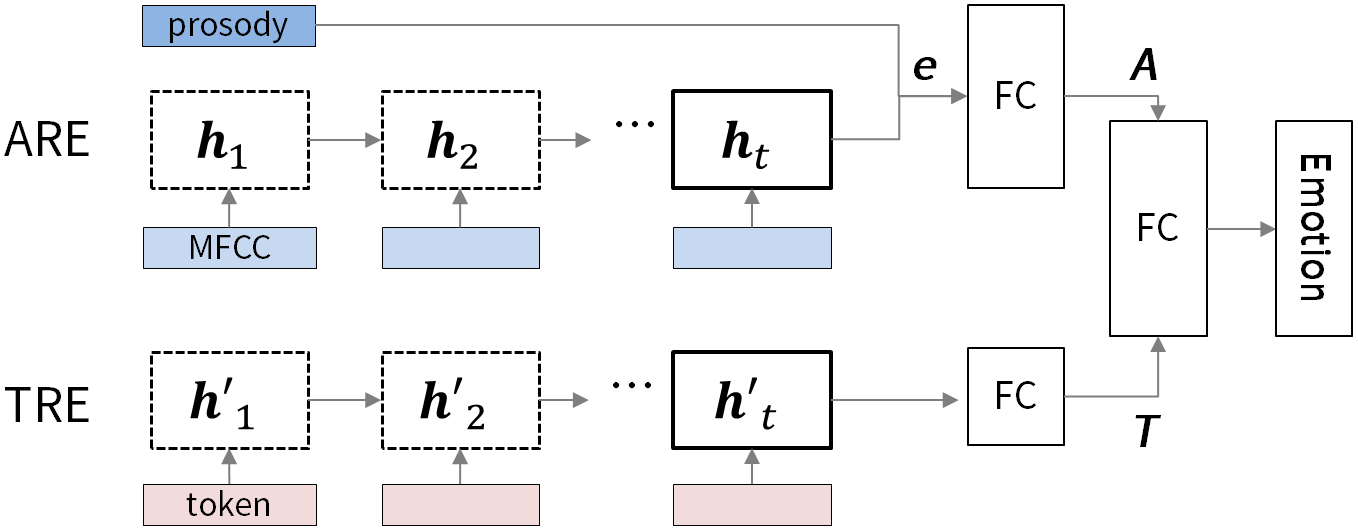}
\caption{
Multimodal dual recurrent encoder. The upper part shows the ARE, which encodes audio signals, and the lower part shows the TRE, which encodes textual information.
}
\label{fig_MDRE}
\end{figure}

Motivated by the architecture used in~\cite{wang2016audio,mirsamadi2017automatic}, we build an audio recurrent encoder (ARE) to predict the class of a given audio signal. 
Once MFCC features have been extracted from an audio signal, a subset of the sequential features is fed into the RNN (i.e., gated recurrent units (GRUs)), which leads to the formation of the network's internal hidden state $h_{t}$ to model the time series patterns. 
This internal hidden state is updated at each time step with the input data $\textbf{x}_{t}$ and the hidden state of the previous time step $h_{t-1}$ as follows:
\begin{equation}
\begin{aligned}
& \textbf{h}_t = f_{\theta}(\textbf{h}_{t-1}, \textbf{x}_t),
\end{aligned}
\label{eq_RNN}
\end{equation}
where $f_{\theta}$ is the RNN function with weight parameter $\theta$, $\textbf{h}_t$ represents the hidden state at t-\textit{${th}$} time step, and $\textbf{x}_t$ represents the t-\textit{${th}$} MFCC features in~$\boldsymbol{\textbf{x}}={\{x_{1:t_a}\}}$.
After encoding the audio signal $\textbf{x}$ with the RNN, the last hidden state of the RNN, $\textbf{h}_{t_a}$, is considered to be the representative vector that contains all of the sequential audio data. 
This vector is then concatenated with another prosodic feature vector, $\textbf{p}$, to generate a more informative vector representation of the signal, $\textbf{e}=\text{concat}\{\textbf{h}_{t_a}, \textbf{p}\}$.
The MFCC and the prosodic features are extracted from the audio signal using the openSMILE toolkit~\cite{eyben2013recent}, $\textbf{x}_t\in\mathbb{R}^{39}~\text{and}~\textbf{p}\in\mathbb{R}^{35}$, respectively.
Finally, the emotion class is predicted by applying the softmax function to the vector $\textbf{e}$.
For a given audio sample $i$, we assume that $y_i$ is the true label vector, which contains all zeros but contains a one at the correct class, and $\hat{y}_i$ is the predicted probability distribution from the softmax layer.
The training objective then takes the following form:

\begin{equation}
\begin{aligned}
\hat{y}_{i} = \text{softmax}(\textbf{e}^\intercal M+b), \\
\mathcal{L} = -\log \prod_{i=1}^{N} \sum_{c=1}^{C} y_{i,c} \text{log} (\hat{y}_{i,c}),
\end{aligned}
\label{eq_ARE_loss}
\end{equation}
where $\textbf{e}$ is the calculated representative vector of the audio signal with dimensionality $e\in\mathbb{R}^{d}$.
The $M \in \mathbb{R}^{d \times C}$ and the bias $b$ are learned model parameters. C is the total number of classes, and N is the total number of samples used in training.
The upper part of Figure~\ref{fig_MDRE} shows the architecture of the ARE model.

\subsection{Text Recurrent Encoder (TRE)}
We assume that speech transcripts can be extracted from audio signals with high accuracy, given the advancement of ASR technologies~\cite{yu2016automatic}.
We attempt to use the processed textual information as another modality in predicting the emotion class of a given signal.
To use textual information, a speech transcript is tokenized and indexed into a sequence of tokens using the Natural Language Toolkit (NLTK)~\cite{bird2004nltk}. 
Each token is then passed through a word-embedding layer that converts a word index to a corresponding 300-dimensional vector that contains additional contextual meaning between words. The sequence of embedded tokens is fed into a text recurrent encoder (TRE) in such a way that the audio MFCC features are encoded using the ARE represented by equation~\ref{eq_RNN}.
In this case, $\textbf{x}_t$ is the t-\textit{${th}$} embedded token from the text input.
Finally, the emotion class is predicted from the last hidden state of the text-RNN using the softmax function.

We use the same training objective as the ARE model, and the predicted probability distribution for the target class is as follows:
\begin{equation}
\begin{aligned}
\hat{y}_{i} = \text{softmax}({\textbf{h}_{\text{last}}}^\intercal M+b), \\
\end{aligned}
\label{eq_TRE_loss}
\end{equation}
where ${\textbf{h}_{\text{last}}}$ is last hidden state of the text-RNN,~${\textbf{h}_{\text{last}}}\in\mathbb{R}^{d}$, and the $M \in \mathbb{R}^{d \times C}$ and bias $b$ are learned model parameters.
The lower part of Figure~\ref{fig_MDRE} indicates the architecture of the TRE model.

\subsection{Multimodal Dual Recurrent Encoder (MDRE)}
We present a novel architecture called the multimodal dual recurrent encoder (MDRE) to overcome the limitations of existing approaches. In this study, we consider multiple modalities, such as MFCC features, prosodic features and transcripts, which contain sequential audio information, statistical audio information and textual information, respectively. These types of data are the same as those used in the ARE and TRE cases.
The MDRE model employs two RNNs to encode data from the audio signal and textual inputs independently. The audio-RNN encodes MFCC features from the audio signal using equation~\ref{eq_RNN}.
The last hidden state of the audio-RNN is concatenated with the prosodic features to form the final vector representation $\textbf{e}$, and this vector is then passed through a fully connected neural network layer to form the audio encoding vector~\textbf{A}.
On the other hand, the text-RNN encodes the word sequence of the transcript using equation~\ref{eq_RNN}.
The final hidden states of the text-RNN are also passed through another fully connected neural network layer to form a textual encoding vector~\textbf{T}.
Finally, the emotion class is predicted by applying the softmax function to the concatenation of the vectors \textbf{A} and \textbf{T}. 
We use the same training objective as the ARE model, and the predicted probability distribution for the target class is as follows:
\begin{equation}
\begin{aligned}
\textbf{A}=g_{\theta}(\textbf{e}),~\textbf{T}={g\textprime}_{\theta}(\textbf{h}_{\text{last}}), \\
\hat{y}_{i} = \text{softmax}(\text{concat}(\textbf{A},\textbf{T})^\intercal M+b), \\
\end{aligned}
\label{eq_MERE_loss}
\end{equation}
where $g_{\theta},{g\textprime}_{\theta}$ is the feed-forward neural network with weight parameter $\theta$, and $\textbf{A}$, $\textbf{T}$ are final encoding vectors from the audio-RNN and text-RNN, respectively. $M \in \mathbb{R}^{d \times C}$ and the bias $b$ are learned model parameters.

\begin{figure}[t]
\small
\centering
\includegraphics[width=1.0\columnwidth]{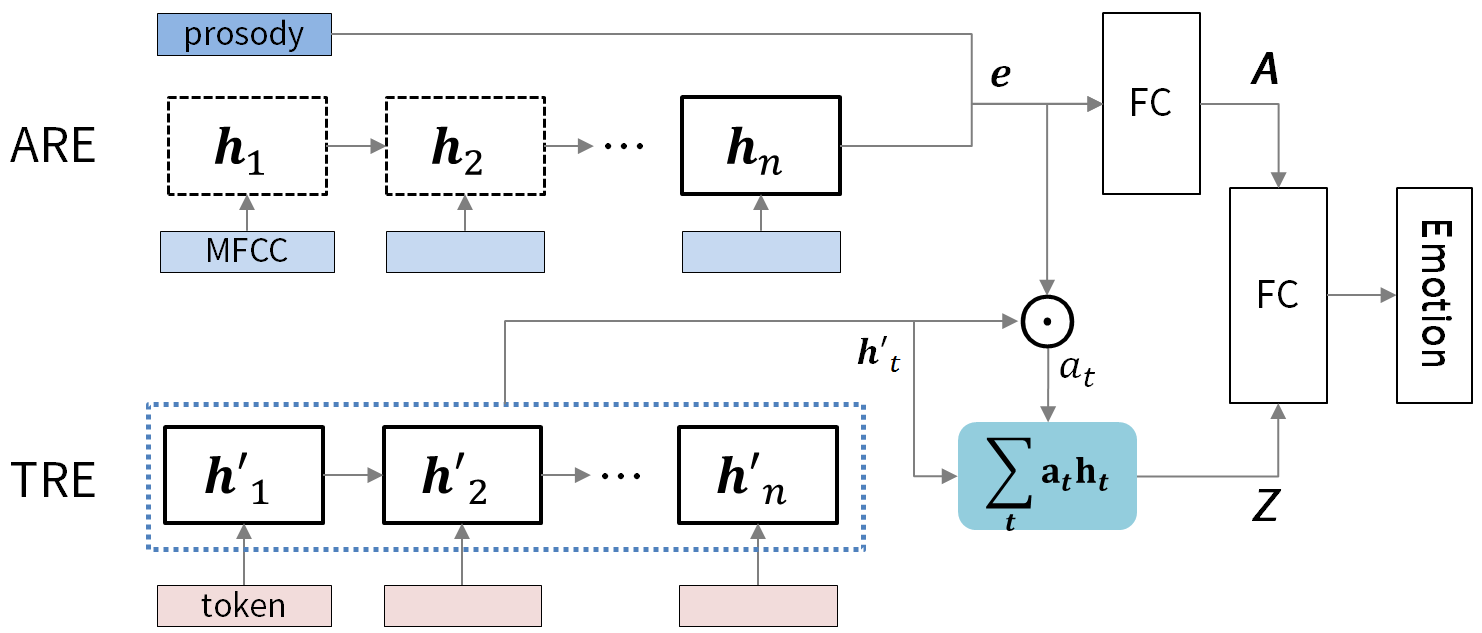}
\caption{
Architecture of the MDREA model. The weighted sum of the sequence of the hidden states of the text-RNN $\textbf{h}_t$ is taken using the attention weight $a_t$; $a_t$ is calculated as the dot product of the final encoding vector of the audio-RNN $\textbf{e}$ and $\textbf{h}_t$.
}
\label{fig_MDREA}
\end{figure}

\subsection{Multimodal Dual Recurrent Encoder with Attention (MDREA)}
Inspired by the concept of the attention mechanism used in neural machine translation~\cite{luong2015effective}, we propose a novel multimodal attention method to focus on the specific parts of a transcript that contain strong emotional information, conditioning on the audio information.
Figure~\ref{fig_MDREA} shows the architecture of the MDREA model.
First, the audio data and text data are encoded with the audio-RNN and text-RNN using equation~\ref{eq_RNN}. 
We then consider the final audio encoding vector $\textbf{e}$ as a context vector. 
As seen in equation~\ref{eq_attn}, during each time step \textit{t}, the dot product between the context vector  \textbf{e} and the hidden state of the text-RNN at each t-\textit{th} sequence $\textbf{h}_t$ is evaluated to calculate a similarity score $a_t$.
Using this score $a_t$ as a weight parameter, the weighted sum of the sequences of the hidden state of the text-RNN,~$\textbf{h}_t$, is calculated to generate an attention-application vector \textbf{Z}.
This attention-application vector is concatenated with the final encoding vector of the audio-RNN $\textbf{A}$ (equation~\ref{eq_MERE_loss}), which will be passed through the softmax function to predict the emotion class.
We use the same training objective as the ARE model, and the predicted probability distribution for the target class is as follows: 
\begin{equation}
\begin{aligned}
	&a_t=\dfrac{\text{exp}(\textbf{e}^\intercal\textbf{h}_{t})}{\sum_t \text{exp}(\textbf{e}^\intercal\textbf{h}_{t})},
   ~\textbf{Z}=\sum_t a_t{\textbf{h}_{t}}, \\
   &\hat{y}_{i,j} = \text{softmax}(\text{concat}(\textbf{Z},\textbf{A})^\intercal M+b), \\
\end{aligned}
\label{eq_attn}
\end{equation}
where $M \in \mathbb{R}^{d \times C}$ and the bias $b$ are learned model parameters.

\section{Experimental Setup and Dataset}
\subsection{Dataset}
We evaluate our model using the Interactive Emotional Dyadic Motion Capture (IEMOCAP)~\cite{busso2008iemocap} dataset. 
This dataset was collected following theatrical theory in order to simulate natural dyadic interactions between actors.
We use categorical evaluations with majority agreement. We use only four emotional categories {\textit{happy}, \textit{sad}, \textit{angry}, and \textit{neutral}} to compare the performance of our model with other research using the same categories.
The IEMOCAP dataset includes five sessions, and each session contains utterances from two speakers (one male and one female). This data collection process resulted in 10 unique speakers. For consistent comparison with previous work, we merge the excitement dataset with the happiness dataset. The final dataset contains a total of 5531 utterances (1636 \textit{happy}, 1084 \textit{sad}, 1103 \textit{angry}, 1708 \textit{neutral}).


\begin{figure*}[b!]
\centering
\subfigure[ARE]{\label{fig_confusion_ARE}\includegraphics[width=0.68\columnwidth]{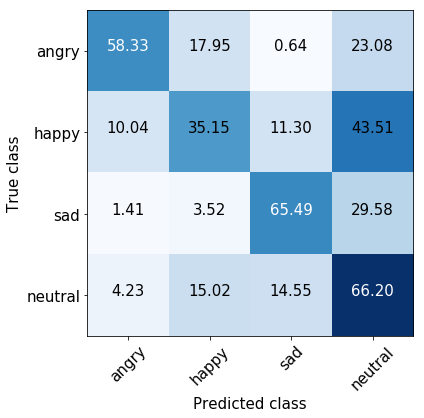}}
\subfigure[TRE]{\label{fig_confusion_TRE}\includegraphics[width=0.68\columnwidth]{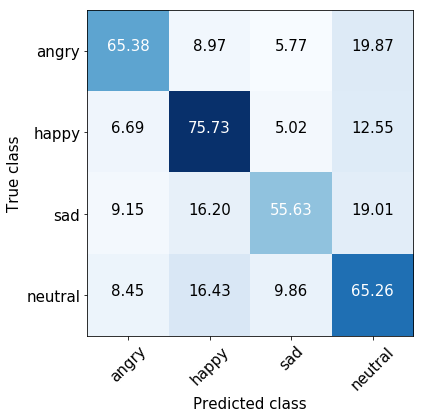}}
\subfigure[MDRE]{\label{fig_confusion_MDRE}\includegraphics[width=0.68\columnwidth]{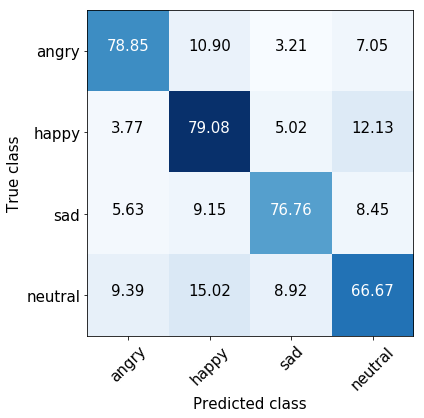}}
\caption{
Confusion matrix of each model.
}
\label{fig_confusion}
\end{figure*}

\subsection{Feature extraction}
To extract speech information from audio signals, we use MFCC values, which are widely used in analyzing audio signals. 
The MFCC feature set contains a total of 39 features, which include 12 MFCC parameters (1-12) from the 26 Mel-frequency bands and log-energy parameters, 13 delta and 13 acceleration coefficients
The frame size is set to 25 ms at a rate of 10 ms with the Hamming function. 
According to the length of each wave file, the sequential step of the MFCC features is varied. 
To extract additional information from the data, we also use prosodic features, which show effectiveness in affective computing. 
The prosodic features are composed of 35 features, which include the F0 frequency, the voicing probability, and the loudness contours. 
All of these MFCC and prosodic features are extracted from the data using the OpenSMILE toolkit~\cite{eyben2013recent}.

\subsection{Implementation details}
Among the variants of the RNN function, we use GRUs as they yield comparable performance to that of the LSTM and include a smaller number of weight parameters~\cite{chung2014empirical}. 
We use a max encoder step of 750 for the audio input, based on the implementation choices presented in~\cite{neumann2017attentive} and 128 for the text input because it covers the maximum length of the transcripts.
The vocabulary size of the dataset is 3,747, including the ``\_UNK\_" token, which represents unknown words, and the ``\_PAD\_" token, which is used to indicate padding information added while preparing mini-batch data. The number of hidden units and the number of layers in the RNN for each model (ARE, TRE, MDRE and MDREA) are selected based on extensive hyperparameter search experiments. 
The weights of the hidden units are initialized using orthogonal weights \cite{saxe2013exact}], and the text embedding layer is initialized from pretrained word-embedding vectors~\cite{pennington2014glove}.

In preparing the textual dataset, we first use the released transcripts of the IEMOCAP dataset for simplicity. To investigate the practical performance, we then process all of the IEMOCAP audio data using an ASR system (the Google Cloud Speech API) and retrieve the transcripts. The performance of the Google ASR system is reflected by its word error rate (WER) of 5.53\%.


\begin{table}[t]
\centering

\begin{tabular}
{C{0.50\columnwidth}C{0.30\columnwidth}}
\hline 
\textbf{Model}\Tstrut & \textbf{WAP}\Tstrut \\ \hline
ACNN~\cite{neumann2017attentive}\Tstrut  & $ 0.561\Tstrut $ \\
LLD RNN-attn~\cite{mirsamadi2017automatic} 	  & $0.635 $  \\
RNN(prop.)-ELM~\cite{lee2015high} 	  & $0.628 $  \\
3CNN-LSTM10H~\cite{satt2017efficient} & $0.688 $  \\

\hline
ARE\Tstrut 	& $0.546~\small{\pm{0.009}}\Tstrut $ \\
TRE 		& $0.635~\small{\pm{0.018}} $  \\
MDRE 		& $\textbf{0.718}~\small{\pm{\textbf{0.019}}} $  \\
MDREA 		& $0.690~\small{\pm{0.019}} $  \\

\hline
TRE-ASR\Tstrut 	& $0.593~\small{\pm{0.022}}\Tstrut $  \\
MDRE-ASR 		& $\textbf{0.691}~\small{\pm{\textbf{0.019}}} $  \\
MDREA-ASR		& $0.677~\small{\pm{0.013}} $  \\

\hline 

\end{tabular}
\caption{
Model performance comparisons. The top 2 best-performing models (according to the unweighted average recall) are marked in bold. The ``-ASR'' models are trained with processed transcripts from the Google Cloud Speech API.
}
\label{table_performance}
\end{table}

\section{Empirical Results}
\subsection{Performance evaluation}
As the dataset is not explicitly split beforehand into training, development, and testing sets, we perform 5-fold cross validation to determine the overall performance of the model. The data in each fold are split into training, development, and testing datasets (8:0.5:1.5, respectively).
After training the model, we measure the weighted average precision (WAP) over the 5-fold dataset. We train and evaluate the model 10 times per fold, and the model performance is assessed in terms of the mean score and standard deviation.

We examine the WAP values, which are shown in Table 1. First, our ARE model shows the baseline performance because we use minimal audio features, such as the MFCC and prosodic features with simple architectures. On the other hand, the TRE model shows higher performance gain compared to the ARE. From this result, we note that textual data are informative in emotion prediction tasks, and the recurrent encoder model is effective in understanding these types of sequential data. Second, the newly proposed model, MDRE, shows a substantial performance gain. It thus achieves the state-of-the-art performance with a WAP value of 0.718. This result shows that multimodal information is a key factor in affective computing. Lastly, the attention model, MDREA, also outperforms the best existing research results (WAP 0.690 to 0.688)~\cite{satt2017efficient}.
However, the MDREA model does not match the performance of the MDRE model, even though it utilizes a more complex architecture. We believe that this result arises because insufficient data are available to properly determine the complex model parameters in the MDREA model. Moreover, we presume that this model will show better performance when the audio signals are aligned with the textual sequence while applying the attention mechanism. We leave the implementation of this point as a future research direction.

To investigate the practical performance of the proposed models, we conduct further experiments with the ASR-processed transcript data (see ``-ASR'' models in Table~\ref{table_performance}). 
The label accuracy of the processed transcripts is 5.53\% WER. The TRE-ASR, MDRE-ASR and MDREA-ASR models reflect degraded performance compared to that of the TRE, MDRE and MDREA models. However, the performance of these models is still competitive; in particular, the MDRE-ASR model outperforms the previous best-performing model, 3CNN-LSTM10H (WAP 0.691 to 0.688).

\subsection{Error analysis}
We analyze the predictions of the ARE, TRE, and MDRE models. 
Figure~\ref{fig_confusion} shows the confusion matrix of each model. The ARE model (Fig.~\ref{fig_confusion_ARE}) incorrectly classifies most instances of \textit{happy} as \textit{neutral} (43.51\%); thus, it shows reduced accuracy (35.15\%) in predicting the the \textit{happy} class.
Overall, most of the emotion classes are frequently confused with the \textit{neutral} class.
This observation is in line with the findings of~\cite{neumann2017attentive}, who noted that the neutral class is located in the center of the activation-valence space, complicating its discrimination from the other classes.

Interestingly, the TRE model (Fig.~\ref{fig_confusion_TRE}) shows greater prediction gains in predicting the \textit{happy} class when compared to the ARE model (35.15\% to 75.73\%).
This result seems plausible because the model can benefit from the differences among the distributions of words in \textit{happy} and \textit{neutral} expressions, which gives more emotional information to the model than that of the audio signal data.
On the other hand, it is striking that the TRE model incorrectly predicts instances of the \textit{sad} class as the \textit{happy} class 16.20\% of the time, even though these emotional states are opposites of one another.

The MDRE model (Fig.~\ref{fig_confusion_MDRE}) compensates for the weaknesses of the previous two models (ARE and TRE) and benefits from their strengths to a surprising degree.
The values arranged along the diagonal axis show that all of the accuracies of the correctly predicted class have increased. Furthermore, the occurrence of the incorrect “\textit{sad}-to-\textit{happy}" cases in the TRE model is reduced from 16.20\% to 9.15\%. 

\section{Conclusions}
In this paper, we propose a novel multimodal dual recurrent encoder model that simultaneously utilizes text data, as well as audio signals, to permit the better understanding of speech data. Our model encodes the information from audio and text sequences using dual RNNs and then combines the information from these sources using a feed-forward neural model to predict the emotion class. Extensive experiments show that our proposed model outperforms other state-of-the-art methods in classifying the four emotion categories, and accuracies ranging from 68.8\% to 71.8\% are obtained when the model is applied to the IEMOCAP dataset. In particular, it resolves the issue in which predictions frequently incorrectly yield the neutral class, as occurs in previous models that focus on audio features.

In the future work, we aim to extend the modalities to audio, text and video inputs. Furthermore, we plan to investigate the application of the attention mechanism to data derived from multiple modalities. This approach seems likely to uncover enhanced learning schemes that will increase performance in both speech emotion recognition and other multimodal classification tasks.

\section*{Acknowledgments}
K. Jung is with the Department of Electrical and Computer Engineering, ASRI, Seoul National University, Seoul, Korea. This work was supported by the Ministry of Trade, Industry \& Energy (MOTIE, Korea) under Industrial Technology Innovation Program (No.10073144).

\bibliographystyle{IEEEbib}
\bibliography{mybib}




\end{document}